\documentclass[conference]{IEEEtran}
\IEEEoverridecommandlockouts
\usepackage{cite}
\usepackage{amsmath,amssymb,amsfonts}
\usepackage{algorithmic}
\usepackage{graphicx}
\usepackage{textcomp}
\usepackage{xcolor}
\usepackage{algorithm}
\usepackage{algorithmic}
\usepackage{amsmath}
\usepackage{amssymb}
\usepackage{bbding}
\usepackage{hyperref}
\hypersetup{colorlinks=true,
	linkcolor=black,
	anchorcolor=black,
	citecolor=black}
\usepackage{booktabs}
\usepackage{multirow}
\def\BibTeX{{\rm B\kern-.05em{\sc i\kern-.025em b}\kern-.08em
    T\kern-.1667em\lower.7ex\hbox{E}\kern-.125emX}}
\begin{document}

\title{Video Anomaly Detection with Semantics-Aware Information Bottleneck}

\author{
  Juntong Li\textsuperscript{*},
  Lingwei Dang\textsuperscript{*}\thanks{\textsuperscript{*}Equal contributions. Email: qlzjftm@gmail.com, levondang@163.com.},
  Qingxin Xiao,  \\
  Shishuo Shang,
  Jiajia Cheng,
  Haomin Wu,
  Yun Hao,
  Qingyao Wu\textsuperscript{\textdagger}\thanks{\textsuperscript{\textdagger}Email: qyw@scut.edu.cn.}\\
  School of Software Engineering, South China University of Technology
   
}

\maketitle

\begin{abstract}

Semi-supervised video anomaly detection methods face two critical challenges: (1) Strong generalization blurs the boundary between normal and abnormal patterns. Although existing approaches attempt to alleviate this issue using memory modules, their rigid prototype-matching process limits adaptability to diverse scenarios; (2) Relying solely on low-level appearance and motion cues makes it difficult to perceive high-level semantic anomalies in complex scenes. To address these limitations, we propose SIB-VAD, a novel framework based on adaptive information bottleneck filtering and semantic-aware enhancement. We propose the Sparse Feature Filtering Module (SFFM) to replace traditional memory modules. It compresses normal features directly into a low-dimensional manifold based on the information bottleneck principle and uses an adaptive routing mechanism to dynamically select the most suitable normal bottleneck subspace. Trained only on normal data, SFFMs only learn normal low-dimensional manifolds, while abnormal features deviate and are effectively filtered. Unlike memory modules, SFFM directly removes abnormal information and adaptively handles scene variations. To improve semantic awareness, we further design a multimodal prediction framework that jointly models appearance, motion, and semantics. Through multimodal consistency constraints and joint error computation, it achieves more robust VAD performance. Experimental results validate the effectiveness of our feature filtering paradigm based on semantics-aware information bottleneck. Project page at \href{https://qzfm.github.io/sib_vad_project_page/}{https://qzfm.github.io/sib\_vad\_project\_page/}

\end{abstract}


\section{Introduction}

Video Anomaly Detection (VAD) aims to identify events that deviate from normal behavior patterns and is widely applied in public security surveillance, transportation, and industrial monitoring. Due to the scarcity of abnormal samples, existing methods typically learn normal patterns from large volumes of normal videos. During inference, if an abnormal event occurs, it will exhibit a large reconstruction~\cite{gong2019memorizing, ristea2024self} or prediction~\cite{liu2018future, zhou2025video} error, thereby enabling anomaly detection. However, current approaches still face two key limitations in complex real-world scenarios: the inseparability of anomalies caused by over-generalization and the lack of semantic-level temporal modeling capability.

\begin{figure}
    \centering
    \includegraphics[width=0.95\columnwidth]{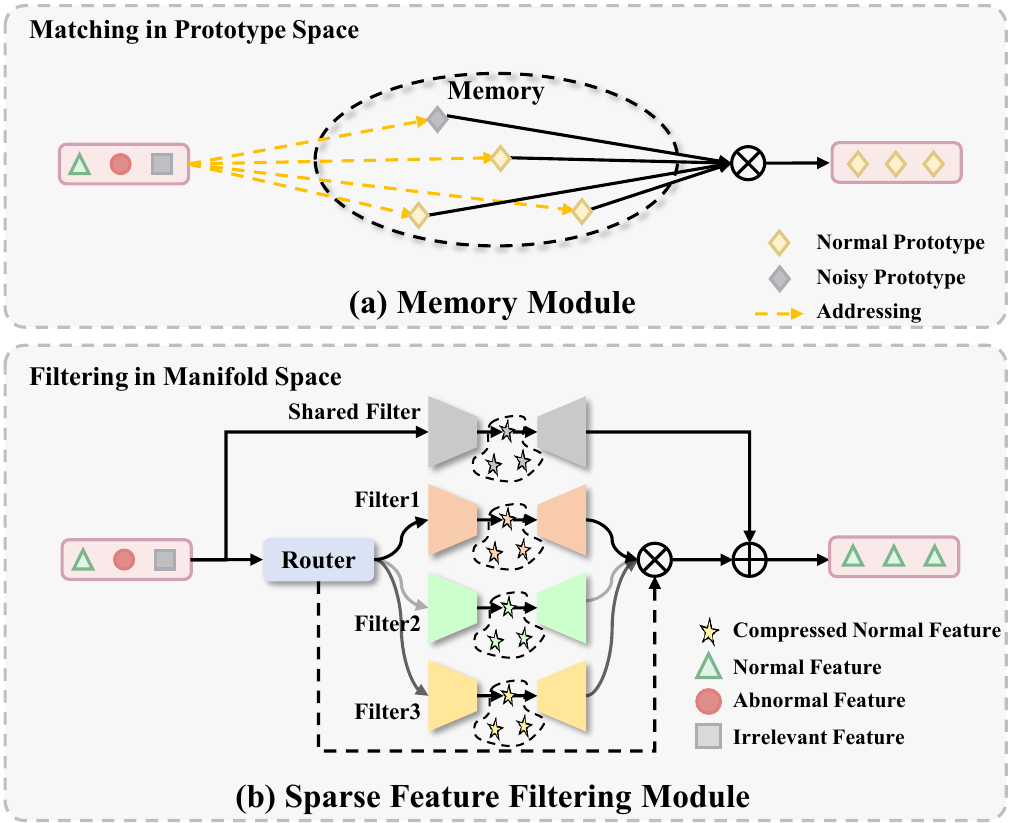}
    \caption{Comparison between the previous and our proposed framework. (a) Existing memory modules filter abnormal information through inflexible prototype matching. (b) Our proposed sparse feature filtering paradigm achieves better VAD performance compared to memory-based methods.}
    \label{fig:top}
\end{figure}

First, the powerful generalization ability of deep neural networks undermines anomaly separability~\cite{gong2019memorizing}. Specifically, models may produce plausible reconstructions or predictions for abnormal events, resulting in overlapping error distributions between normal and abnormal data. To mitigate this, several works~\cite{gong2019memorizing,liu2021hf2vad, lyu2025vadmamba} introduce memory modules (Fig. \ref{fig:top}(a)) to memorize normal prototypes. However, memory-based methods have intrinsic limitations: a fixed number of prototypes cannot capture the full diversity of normal patterns, and rigid prototype-matching fails to adapt to variations across scenes and events. DLAN-AC~\cite{yang2022dynamic} attempts to address these by determining memory slots for different datasets via pre-clustering, and MA-PDM~\cite{zhou2025video} increases memory capacity with fine-grained patch memories. Despite this, due to the fixed size and rigid structure of memory modules, these strategies remain unable to effectively represent all normal patterns or adapt to scene variations.

\begin{figure*}
    \centering
    \includegraphics[width=0.9\textwidth]{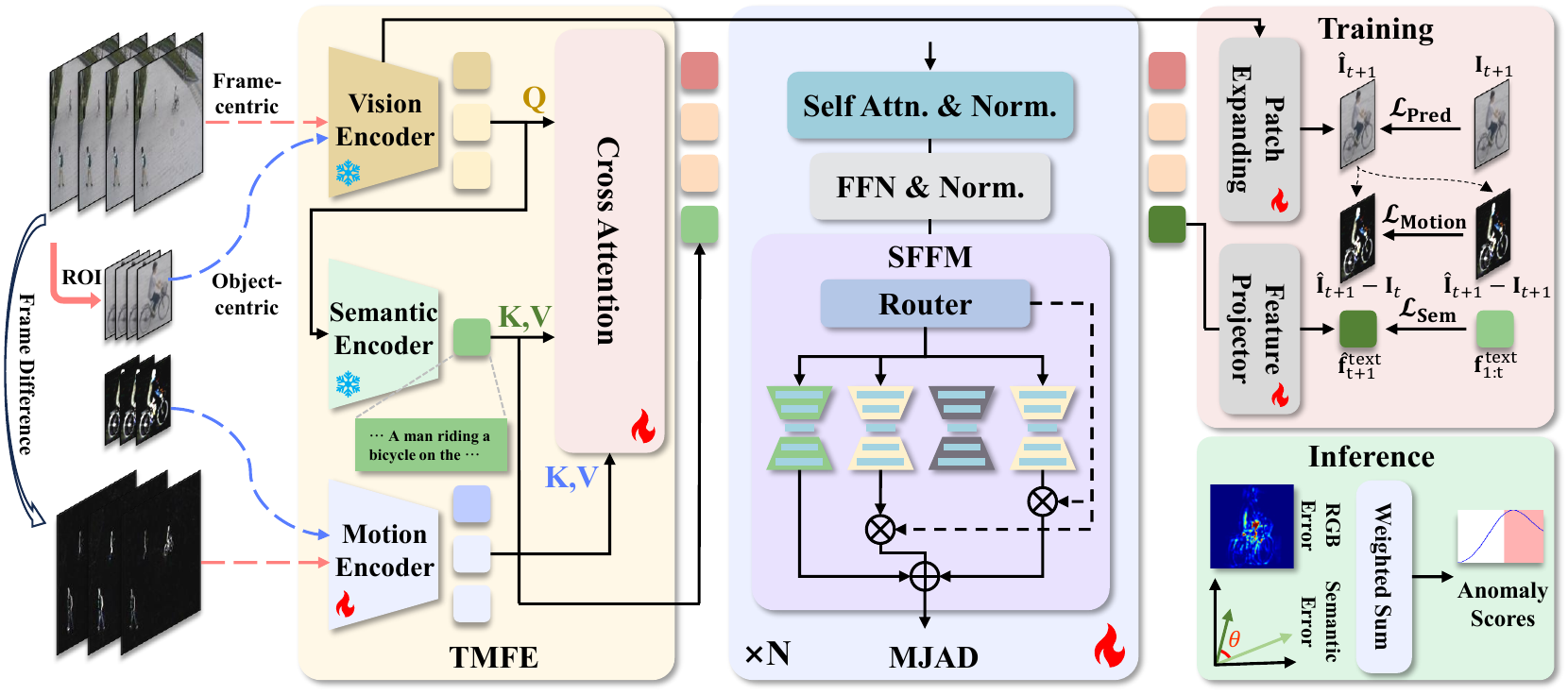}
    \caption{Overview of SIB-VAD. Firstly, TMFE (Sec. \ref{tmfe}) extracts descriptions from the input clips and encodes and fuses semantic, appearance, and motion features. Secondly, MJAD (Sec. \ref {mmjd}) performs joint decoding of the fused features to predict the next frame and semantic features. SFFM (Sec. \ref{sffm}) filters abnormal information to increase the prediction error when anomalies occur. Finally, anomaly scores are calculated based on frame prediction errors and semantic errors (Sec. \ref {anomaly_detection}).}
    \label{fig:framework}
\end{figure*}


Second, relying solely on low-level appearance~\cite{liu2018future,zhou2025video} and motion~\cite{liu2018future,lyu2025vadmamba} cues makes it difficult to capture subtle semantic anomalies from RGB errors. Although some works~\cite{zhang2024holmesvau, wang2025learning, 10687724} attempt to leverage the strong semantic extraction capabilities of Vision-Language Models (VLMs), these models either can only produce shallow textual descriptions and fail to achieve frame-level detection, or they treat text as static external prior supervision, constraining visual features through semantic labels rather than enabling dynamic semantic awareness.


To address the above limitations, we propose a novel VAD framework, termed \textbf{SIB-VAD} (Fig. \ref{fig:framework}), based on adaptive information bottleneck filtering and semantic-aware enhancement.

The core insight is that enhancing anomaly separability does not lie in ``memorizing normality'', but on more effectively and adaptively compressing features and filtering abnormal information to prevent anomalies from being well reconstructed or predicted. Inspired by the information bottleneck principle~\cite{tishby2000information}, which leverages bottleneck structures to balance representation capacity and irrelevant information compression. We design the bottleneck filters to directly compress features. Trained only on normal data, these compressed features are forced to lie on a low-dimensional manifold corresponding to normal patterns, while abnormal information falls outside this space. To further model the diversity of normal patterns and achieve adaptive filtering, we propose a \textbf{Sparse Feature Filtering Module (SFFM, Fig. \ref{fig:top} (b))}, which integrates multiple filters and dynamically selects subspaces through an adaptive routing mechanism, followed by weighted aggregation. Unlike memory modules that rely on rigid prototype matching, the adaptive bottleneck filtering process is more direct and flexible. Notably, abnormal information can also misroute features, further amplifying prediction differences, which helps distinguish anomalies more effectively.

To introduce fine-grained semantic modeling, we construct a multimodal framework. By incorporating semantic descriptions, we jointly model appearance, motion, and semantics. The SFFMs are integrated into the decoder to enable effective filtering of abnormal information, ultimately predicting both future frames and semantic features. This allows semantic errors to serve as an additional signal for VAD.

Furthermore, we introduce a multimodal consistency enhancement mechanism, which enforces consistency across visual, motion, and semantic predictions through frame prediction loss, semantic similarity loss, and motion frame-difference contrastive loss. In particular, the motion frame-difference contrastive loss not only improves sensitivity to motion patterns but also suppresses shortcut predictions, promoting faithful modeling of real temporal dynamics. Finally, visual and semantic prediction errors are jointly used to compute anomaly scores, achieving more robust anomaly detection.

Our contributions are summarized as follows:
\begin{itemize}
    \item We propose a novel abnormal information filtering paradigm, combining the information bottleneck principle with an adaptive routing mechanism to construct a continuous and flexible filtering structure, effectively mitigating the over-generalization problem in existing methods.
    
    \item We design a multimodal joint modeling framework that captures appearance, motion, and semantic information simultaneously, leveraging cross-modal cues to enhance sensitivity to diverse anomalies.
    
    \item We introduce a multimodal consistency enhancement mechanism, which enforces cross-modal consistency and jointly models prediction errors to achieve more robust anomaly scoring and detection performance.
\end{itemize}


\section{Related Work}

\textbf{Video Anomaly Detection.} VAD aims to identify events that deviate from normal behavior patterns. Existing VAD methods typically adopt a semi-supervised approach, learning normal patterns from large volumes of normal videos and detecting anomalies based on significant reconstruction~\cite{gong2019memorizing, ristea2024self} or prediction~\cite{liu2018future, zhou2025video} errors during inference.

\textbf{Memory networks.} To prevent overgeneralization to anomalies, prior works~\cite{gong2019memorizing,liu2021hf2vad, lyu2025vadmamba, zhou2025video, yang2022dynamic} use memory modules to store normal patterns. However, fixed prototypes limit feature diversity and ignore scenario differences. Inspired by the information bottleneck principle, we designed the SFFM for more direct and flexible filtering. 

\textbf{Vision-Language Models.} VLMs~\cite{radford2021learning, chen2024expanding} possess powerful semantic reasoning capabilities. Existing VLM-based VAD methods~\cite{zhang2024holmesvau,10687724} can only produce textual descriptions and lack awareness of frame-level fine-grained information, or just treat text as a static prior. We integrated VLMs into a fine-grained multimodal prediction framework to comprehensively capture the semantic cues of anomalies.


\section{Method}
The framework we propose, termed \textbf{SIB-VAD}, is illustrated in Figure \ref{fig:framework}. First, \textbf{Tri-Modal Fusion Encoder (TMFE, Sec. \ref{tmfe})}  encodes frames, frame differences, and semantic descriptions to a joint model multimodal information. The fusion features are then passed to the \textbf{Multimodal Joint Attention Decoder (MJAD, Sec. \ref{mmjd})} for mutual assistance modeling. Each layer incorporates a \textbf{Sparse Feature Filtering Module (SFFM, Sec. \ref{sffm})} to filter abnormal information. Finally, we introduce the Inter-modality Consistency Constraint Mechanism to ensure the multimodal alignment. We will first elaborate on the problem formulation (Sec. \ref{problem_definition}), then provide detailed explanations of three modules, and finally show how to implement anomaly detection (Sec. \ref{anomaly_detection}).

\subsection{Problem Definition}
\label{problem_definition}
Following~\cite{liu2018future}, we employ a prediction-based method where a video prediction model is trained on normal data. Mathematically, given consecutive video frames $\{\mathbf{I}_1,\mathbf{I}_2,\cdots,\mathbf{I}_t\}$, the model predicts the next frame $\mathbf{I}_{t+1}$ and its corresponding semantic feature $\mathbf{f}_{t+1}^{\text{text}}$. During the training phase, the model is trained using only normal samples to perform a prediction task. During testing, both frame and semantic errors are used as the frame-level anomaly score. Notably, our framework supports both frame-centric and object-centric paradigms: the former predicts the entire frame, while the latter processes each individual object in the scene as an independent unit.

\subsection{Tri-Modal Fusion Encoder}
\label{tmfe}
Some works \cite{zhang2024holmesvau, wang2025learning, 10687724} have attempted to apply VLMs to VAD tasks; however, most existing semi-supervised VAD methods focus on appearance and motion, overlooking the high-level semantics of anomalies. For joint modeling, we employ a lightweight VLM~\cite{chen2024expanding} to extract video descriptions of input video clips. Specifically, we first utilize the vision encoder from the pre-trained VLM to extract visual embeddings $\mathbf{F}_{i}^{h}$ for continuous video frames $\{\mathbf{I}_i\}$, where $i=1,2,\cdots,t$. 

With the visual features, a lightweight language model~\cite{yang2024qwen2} generates the semantic description $C_{1:t}$ for the input clip. We employ a pre-trained sentence encoder~\cite{gao2021simcse} to extract global semantic embeddings $\mathbf{f}_{1:t}^\text{text}$. 

Previous work \cite{liu2018future, liu2021hf2vad} has demonstrated that motion information plays a crucial role in the VAD task. Unlike these methods that rely on computationally derived optical flow to represent motion, we use frame differences ($\mathbf{D}_i = \mathbf{I}_i - \mathbf{I}_{i-1},  \; i=2,3,\cdots,t$) to directly capture the raw temporal evolution with low computational complexity. We train a ViT-based motion encoder to effectively extract motion embeddings $\mathbf{F}_{i}^{m},$ where $i=2,3,\cdots,t.$

To enable deeper interaction between triple modals, we use a cross-attention module to inject semantic and motion information into visual features:  
\begin{equation}
    \mathbf{F}^{\text{fusion}} = \text{CrossAttn}({\mathbf{F}}^{h}, [\mathbf{F}_{1:t}^{\text{text}}, \mathbf{F}^{m}]) + \mathbf{F}^{h}.
\end{equation}

Then we insert the semantic embeddings $\mathbf{f}_{1:t}^{\text{text}}$ into the fusion token sequence. In the subsequent joint decoder, we simultaneously predict future frames and semantics.

\subsection{Sparse Feature Filtering Module}
\label{sffm}
\paragraph{Bottleneck Feature Filter}
Our goal is to achieve direct filtering of abnormal information rather than explicitly memorizing normal prototypes. Inspired by the information bottleneck principle~\cite{tishby2000information}, we designed a simple yet effective bottleneck feature filter. These filters learn to extract low-dimensional key information by training on normal data to perform prediction tasks. Since they only learn the normal manifold, during the testing phase, abnormal information residing outside the representation space will be filtered out, causing abnormal events to be inaccurately predicted. We define the bottleneck feature filtering process $\text{BFF}$ as follows: 
\begin{equation}
    \mathrm{BFF}(\cdot) = \mathrm{Expand}(\mathrm{Act}(\mathrm{Reduce}(\cdot))),
\end{equation}
where $\mathrm{Expand}(\cdot)$ and $\mathrm{Reduce}(\cdot)$ denote the channel-wise expansion and reduction processes, respectively, and $\mathrm{Act}(\cdot)$ represents the activation function. We use a Multi-Layer Perceptron (MLP) to implement the BFF process.

\paragraph{Adaptive Routing Mechanism}
To model the diversity of normal patterns, horizontally stacking filters is intuitive but ineffective, as it widens the bottleneck. We propose the Sparse Feature Filtering Module (SFFM), an architecture that stacks filters horizontally while using an adaptive mechanism to selectively route features and implement fine-grained filtering. Thanks to the routing process, the possibility of abnormal features being mishandled and thereby also amplifying the prediction error increases. SFFM can be formalized as follows:
\begin{equation}
    \left\{\begin{aligned}
    \mathbf{f}_{\text{out}} &= \sum_{i=1}^{N}\left(g_i \times \mathrm{BFF}_i(\mathbf{f_{in}})\right), \\
    g_{i} &= 
        \begin{cases}
            s_{i}, & s_i \in \mathrm{TopK}\left(\{s_i\}\right), \\
            0, & \text{otherwise}, 
        \end{cases} \\
    s_i &= \mathrm{Softmax}_i(\mathrm{Routing}(\mathbf{f_{in}})), 1 \leqslant i \leqslant N, 
    \end{aligned} \right.
\end{equation}
where $\mathbf{f_{in}}$ is the input feature, $N$ denotes the total number of filters, $\mathrm{BFF}_i$ is the $i\text{-th}$ filter, $g_i$ indicates the routing score for the $i\text{-th}$ filter, $s_i$ denotes the softmax score between the input and the $i\text{-th}$ filter, $\mathrm{Routing}$ denotes the linear routing.

We observe that this adaptive routing mechanism is prone to routing collapse, where scores persistently concentrate on a small subset of filters. This is due to positive feedback in training: frequently selected filters gain more updates and further cement their priority. To prevent this performance degradation, we introduce the balance loss~\cite{dai2024deepseekmoe}:
\begin{equation}
    \left\{\begin{aligned} 
    \mathcal{L}_\text{Balance} &= \sum_{i=1}^{N} {f_i p_i}, \\
    f_i &= \frac{N}{KT} \sum_{t=1}^{T} \mathbf{1}(\mathbf{f}_t \ \text{is routed to Filter}_i), \\
    p_i &= \frac{1}{T} \sum_{t=1}^{T}{s_{i,t}} ,
    \end{aligned} \right.
\end{equation}
where $T$ is the total number of input features and $\mathbf{1}(\cdot)$ denotes the indicator function.

\subsection{Multimodal Joint Attention Decoder}
\label{mmjd}
Multimodal Joint Attention Decoder (MJAD) is a ViT architecture decoder used to predict the next frame and semantic feature, ensuring mutual assistance among the appearance, motion, and semantic modalities, while inserting an SFFM at each layer to perform feature compression and filtering. A patch expanding process is used to restore spatial size and output the predicted frame $\hat{\mathbf{I}}_{t+1}$, along with a feature projector to project semantic embeddings back to the original space, obtaining the predicted semantic features $\hat{\mathbf{f}}_{t+1}^{\text{text}}$.

\subsection{Inter-modality Consistency Constraint Mechanism}
\label{iccm}
Basically, for the video prediction, we use the $\mathcal{\ell}_2$ distance between the predicted frame $\hat{\mathbf{I}}_{t+1}$ and the ground truth $\mathbf{I}_{t+1}$ as the frame prediction loss: $\mathcal{L}_{\text{Pred}} = \| \mathbf{I}_{t+1} - \hat{\mathbf{I}}_{t+1} \|_2^2.$

For the semantic modality, under normal circumstances, the previous continuous $t$-frame clip should be similar to the semantics of the next frame. The semantic similarity loss is as follows: $\mathcal{L}_{\mathrm{Sem}}
= 1 - \cos(\mathbf{f}_{1:t}^{\text{text}},\, \hat{\mathbf{f}}_{t+1}^{\text{text}}).$

We propose a novel motion frame-difference contrastive loss: when the difference between the prediction $\hat{\mathbf{I}}_{t+1}$ and ground truth $\mathbf{I}_{t+1}$ exceeds the difference between $\hat{\mathbf{I}}_{t+1}$ and the previous ground truth $\mathbf{I}_{t}$, we impose a penalty. This design offers two benefits: first, it indirectly enforces motion constraints in predictions. Second, it prevents the model from achieving low losses by merely mimicking the previous frame through skips, which are only used to restore appearance details. Otherwise, the decoder and filters would not work properly. The motion frame-difference contrastive loss is defined as:
\begin{equation}
    \mathcal{L}_{\text{Motion}}=\mathrm{max}(0, \ (\hat{\mathbf{I}}_{t+1}-\mathbf{I}_{t+1})^2-(\hat{\mathbf{I}}_{t+1}-\mathbf{I}_{t})^2).
\end{equation}

Comprehensively, the object function is conducted as:
$ \mathcal{L}=\lambda_{1}\mathcal{L}_{\text{Pred}} + 
        \lambda_{2}\mathcal{L}_{\text{Sem}} +
        \lambda_{3}\mathcal{L}_{\text{Motion}} +
        \lambda_{4}\mathcal{L}_\text{Balance},$
where $\lambda_{1}$,$\lambda_{2}$,$\lambda_{3}$, and $\lambda_{4}$ are hyperparameters.

\subsection{Anomaly Detection}
\label{anomaly_detection}
During the testing phase, we measure anomaly scores based on prediction errors. We define the visual anomaly score $\mathcal{S}_v$ and the semantic anomaly score $\mathcal{S}_t$ as: $\mathcal{S}_v = \mathcal{G}(\| \mathbf{I}_{t+1} - \hat{\mathbf{I}}_{t+1} \|_2^2)$, $\mathcal{S}_t = \mathcal{G}(1 - \cos(\mathbf{f}_{1:t}^{\text{text}},\, \hat{\mathbf{f}}_{t+1}^{\text{text}}))$, where $\mathcal{G}(\cdot)$ denotes the gaussian smoothing operation. We calculate the weighted sum as frame-level anomaly scores: $\mathcal{S} = w_v \mathcal{S}_v + w_t \mathcal{S}_t$, where $w_v$ and $w_t$ are the weights of the two scores, with higher anomaly scores indicating more likely abnormalities.

\begin{table}[t]
    \centering
    \small
    \caption{Comparison with SOTA methods on three datasets. \textbf{Bold} and \underline{underlined} indicate best and second-best performance.}    
    \begin{tabular}{c|c|c|ccc}
    \toprule

    \multirow{2.5}{*}{} & \multirow{2.5}{*}{\textbf{Method}} & \multirow{2.5}{*}{\textbf{Type}} & \multicolumn{3}{c}{\textbf{AUC}} \\

    \cmidrule(l){4-6}
     &  &  & \textbf{SHT} & \textbf{Avenue} & \textbf{Ped2}  \\
    \midrule
    
    \multirow{9}{*}{\rotatebox{270}{Frame-centric}}
    & FFP \cite{liu2018future} & Pred & 72.8 & 84.9 & 95.4 \\ 
    & MemAE\cite{gong2019memorizing} & Recon & 71.2 & 83.3 &  94.1 \\ 
    & DLAN-AC\cite{yang2022dynamic} & Recon & 74.7 & 89.9 & 97.6 \\ 
    & USTN-DSC\cite{yang2023video} & VidRest & 73.8 & 89.9 & 98.1 \\ 
    & PDM-Net\cite{huang2024long} & Pred & 74.2 & 88.1 & 97.7 \\ 
    & AED-MAE\cite{ristea2024self} & Patch recon & 79.1 &  91.3 & 95.4 \\ 
    & MA-PDM\cite{zhou2025video} & Pred & \underline{79.2} & 91.3 & \underline{98.6} \\ 
    & VADMamba\cite{lyu2025vadmamba} & Hybrid & 77.0 & \underline{91.5} & 98.5 \\
    \cmidrule(l){2-6}
    & \textbf{SIB-VAD (Ours)} & Pred & \textbf{83.4} & \textbf{91.6} & \textbf{98.7} \\
    
    \cmidrule(lr){1-6}

    \multirow{8}{*}{\rotatebox{270}{Object-centric}}
    & VEC\cite{yu2020cloze} & Cloze test & 74.8 & 90.2 &  97.3 \\ 
    & AED-SSMTL\cite{georgescu2021anomaly} & Multi-tasks & 82.4 & 91.5 & 97.5 \\ 
    & $\text{HF}^2$-VAD\cite{liu2021hf2vad} & Hybrid & 76.2 & 91.1 & \textbf{99.3} \\ 
    & FBAE\cite{cao2023new} & Pred & 79.2 & 86.8 & - \\ 
    & MPT\cite{shi2023video} & Multi-tasks & 78.8 & 90.9 & 97.6 \\ 
    & SSAE\cite{cao2024scene} & Pred & 80.5 & 90.2 & - \\ 
    & ADSM\cite{zhang2025autoregressive} & Likelihood & \underline{84.5} & \underline{91.6} & - \\
    \cmidrule(l){2-6}
    & \textbf{SIB-VAD (Ours)} & Pred & \textbf{84.7} & \textbf{92.3} &  \underline{98.9} \\
    
    \bottomrule
    \end{tabular}

    \label{cmp_w_sota}
\end{table}

\section{Experiments}
\label{experiments}
\subsection{Experimental Setup}
\paragraph{Datasets}
We validate the performance of the proposed method on the following three datasets:
\begin{itemize}
    \item \textbf{ShanghaiTech}~\cite{liu2018future}: A highly challenging dataset containing 330 training and 130 testing videos, spanning 13 distinct scenes and 130 types of abnormal events, such as chasing, fighting, cycling, etc.
    \item \textbf{CUHK Avenue}~\cite{lu2013abnormal}: Includes 16 training and 21 testing videos, all collected in campus scenarios. Anomaly types involve strange actions, throwing objects, etc.
    \item \textbf{UCSD Ped2}~\cite{li2013anomaly}: Contains 16 training and 12 testing videos, captured in surveillance pedestrian walking areas. The dataset includes anomalies such as non-pedestrian objects (\textit{e.g.}, bicycles, cars) or abnormal pedestrian behaviors (\textit{e.g.}, running, skateboarding).
\end{itemize}

\paragraph{Evaluation Metric}
We adopted the frame-level area under the curve (AUC) as the evaluation metric for the model. The AUC comprehensively evaluates model performance by calculating the area under the receiver operating characteristic (ROC) curve, which reflects the model's capability to distinguish between normal and anomalous frames. 

\paragraph{Implementation Details}
In each SFFM, we default to 1 shared filter and 63 adaptive filters, with the top 7 routing filters activated based on the highest routing scores. Each filter is a two-layer bottleneck MLP. The semantic encoder consists of the InternVL2.5-1B~\cite{chen2024expanding} and the text encoder SimCSE-BERT~\cite{gao2021simcse}.
The learning rate is set to 0.0002. $\lambda_{1}$,$\lambda_{2}$,$\lambda_{3}$ are all configured as 1 and $\lambda_{4}$ is 0.001. The anomaly score weights ($w_v$, $w_t$) for Ped2, Avenue, and ShanghaiTech datasets are set to (0.8, 0.2), (0.9, 0.1), (0.2, 0.8), respectively. Our approach follows a prediction-based paradigm, utilizing $t=4$ consecutive input frames to predict both the 5th frame and its semantic features, and only normal data is available during the training phase. In our object-centric approach, YOLOv11~\cite{khanam2024yolov11} extracts objects for independent processing, and the frame-level anomaly score is the maximum score among detected objects. All experiments are conducted on four NVIDIA GeForce RTX 4090 GPUs. More implementation details and the detailed architecture design can be found in the appendix.

\subsection{Comparison with State-of-the-art}
We compare our method with various SOTA methods on three benchmark datasets, including both \textbf{object-centric}~\cite{yu2020cloze, georgescu2021anomaly, liu2021hf2vad, cao2023new, shi2023video, cao2024scene, zhang2025autoregressive} and \textbf{frame-centric }methods~\cite{liu2018future, gong2019memorizing, park2020learning, yang2022dynamic, yang2023video,  ristea2024self, lyu2025vadmamba, zhou2025video}, covering representative \textbf{memory-based} methods~\cite{liu2021hf2vad, gong2019memorizing, park2020learning, yang2022dynamic, lyu2025vadmamba, zhou2025video}. As shown in the table \ref{cmp_w_sota}, our method outperforms other approaches of the same category in both frame-centric and object-centric settings. Notably, we achieve AUC scores of 83.4 (frame-centric) and 84.7 (object-centric) on the challenging SHT dataset. And the entire frame input alone surpasses most object-centric methods without requiring per-object computation in the scene. These demonstrate the efficacy of our multimodal framework and bottleneck filtering mechanism.
\definecolor{darkgreen}{RGB}{0,150,0}

\begin{table}[!t]
    \centering
    \small
    \caption{Ablation analysis of SFFM strategy and comparison with memory modules on ShanghaiTech.}
    \begin{tabular}{c|cccc|c}
    \toprule
    \multicolumn{5}{c|}{\textbf{Strategy}} & \textbf{AUC} \\
    \midrule
    \multicolumn{5}{c|}{w/o} & 79.5 \\
    \midrule
    \multicolumn{5}{c|}{Memory (128 Slots)} & 80.8 \\
    \multicolumn{5}{c|}{Memory (256 Slots)} & 81.1 \\
    \multicolumn{5}{c|}{Memory (512 Slots)} & 80.9 \\
    \multicolumn{5}{c|}{Memory (1024 Slots)} & 80.4 \\
    \multicolumn{5}{c|}{Multi-Level Memory with 256 Slots} & 70.7 \\
    \midrule
     & $K$ & $N$ & $K_s$ & Reduction Rate &  \\
    \midrule
    Standard MLP & 0 & 0 & 1 & $1/4$ & 80.1 \\
    Single Filter & 0 & 0 & 1 & 2 & 81.9 \\
    \midrule
    SFFM (2/16+0) & 2 & 16 & 0 & 4 & 81.6 \\
    SFFM (1/15+1) & 1 & 15 & 1 & 4 & 82.0 \\
    SFFM (3/31+1) & 3 & 31 & 1 & 8 & 82.3 \\
    SFFM (7/63+1) & 7 & 63 & 1 & 16 & \textbf{83.4} \\
    \bottomrule
    \end{tabular}
    \label{tab:sffm_mem}
\end{table}



\begin{table}[!t]
    \centering
    \small
    \caption{Effectiveness analysis of multimodal branches and loss function components on ShanghaiTech.}
    \begin{tabular}{ccc|c||cccc|c}
    \toprule
    \textbf{V} & \textbf{S}& \textbf{M} & \textbf{AUC} &
    \textbf{$\mathcal{L}_\text{Pred}$} & \textbf{$\mathcal{L}_\text{Sem}$} &
    \textbf{$\mathcal{L}_\text{Motion}$} & \textbf{$\mathcal{L}_\text{Bal.}$} & \textbf{AUC} \\
    
    \midrule
    \checkmark &  &  & 71.3 &
    \checkmark &  &  & \checkmark & 72.5 \\
    
     & \checkmark &  & 79.0 &
    \checkmark &  & \checkmark & \checkmark & 72.6 \\
    
    \checkmark & \checkmark &  & 81.1 &
    \checkmark & \checkmark &  & \checkmark & 81.9 \\
    
    \checkmark &  & \checkmark & 72.3 &
    \checkmark & \checkmark & \checkmark &  & 82.5 \\
    
    \checkmark & \checkmark & \checkmark & \textbf{83.4} &
    \checkmark & \checkmark & \checkmark & \checkmark & \textbf{83.4} \\
    
    \bottomrule
    \end{tabular}
    \label{tab:mmbanch_loss}
\end{table}
\begin{figure}
    \centering
    \includegraphics[width=0.95\columnwidth]{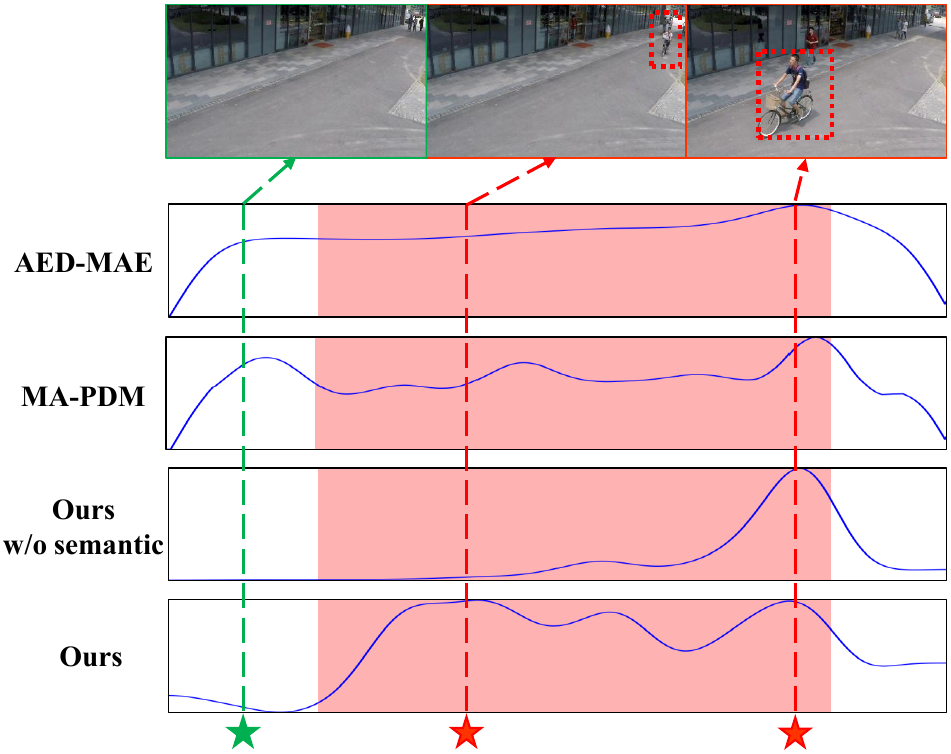}
    \caption{Visualization of anomaly score curves. The images above show the corresponding \textcolor{red}{abnormal} or \textcolor{darkgreen}{normal} events.}
    \label{fig:anomaly_scores}
\end{figure}

\subsection{Ablation Study}
\paragraph{Comparison with Memory Modules}
To demonstrate the superiority of our SFFM over memory modules, we follow prior works~\cite{liu2021hf2vad, gong2019memorizing} by integrating both semantic and visual memory modules at the bottleneck for comparison. As shown in Table \ref{tab:sffm_mem}, while most memory-based methods slightly outperform the baseline, our SFFM achieves significantly better results. This confirms that SFFM enables more direct and effective feature filtering. For fair comparison, we also replaced each SFFM in the decoder with memory modules, but observed a sharp performance drop due to over-filtering.

\paragraph{Impact of SFFM}

We validate the SFFM and explore partitioning strategies. As shown in Table \ref{tab:sffm_mem}, the bottleneck filter significantly outperforms the standard MLP, confirming its effectiveness. Gradually refining the filter partitioning yields the best performance, indicating that more fine-grained partitions enable specialized filtering. Ignoring shared filters causes a slight performance drop, indicating that it is necessary to introduce shared filters to isolate shared information.

\paragraph{Effectiveness of Multimodal Branches}
Table \ref{tab:mmbanch_loss} shows that on ShanghaiTech, single vision or motion branches perform poorly. In contrast, the semantic branch yields 79.0 AUC, rising to 81.1 when fused with vision. Combining all three modalities reaches 83.4, demonstrating the effectiveness of multimodal complementarity.

\paragraph{Effectiveness of Loss Functions}
Table \ref{tab:mmbanch_loss} validates our approach. While the baseline (appearance and motion) yields 72.6, incorporating semantic loss significantly boosts performance to 81.9. Combining all three modalities reaches the best AUC of 83.4. Additionally, the balanced loss proves essential for preventing SFFM degeneration.

\begin{figure}
    \centering
    \includegraphics[width=\columnwidth]{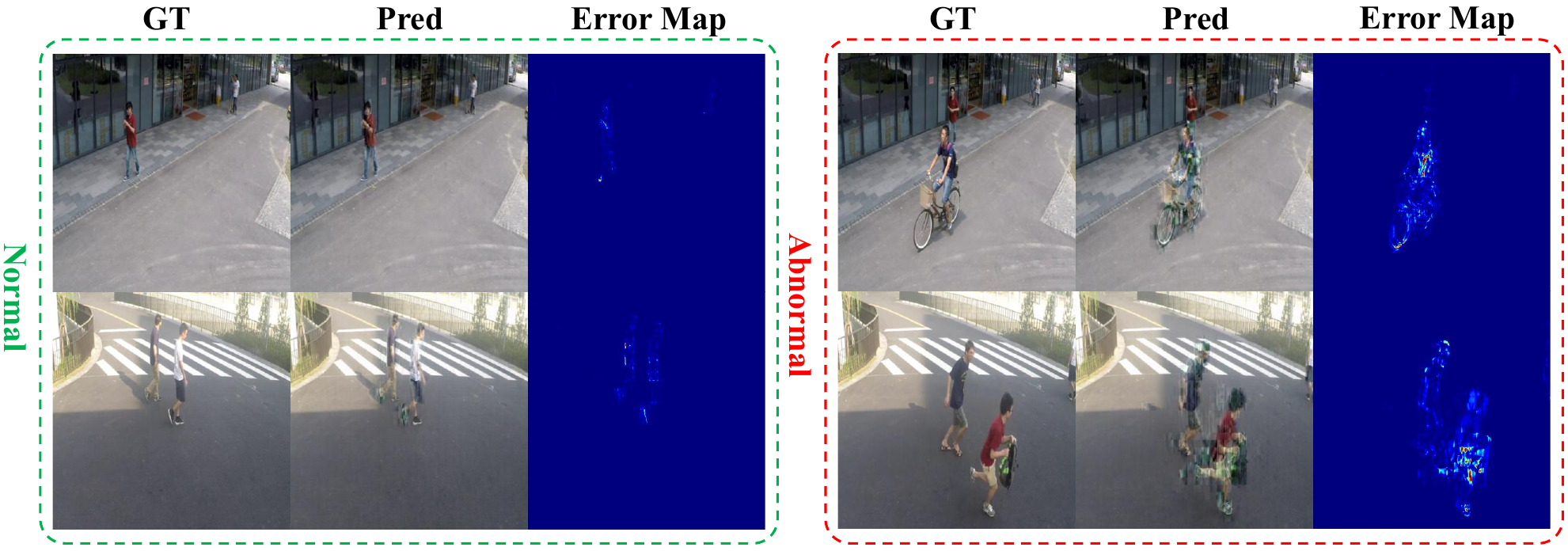}
    \caption{Visualization of frame-centric predictions and error maps. In the error maps, brighter areas indicate higher levels of abnormality in those regions.}
    \label{fig:error_maps}
\end{figure}

\subsection{Visualization Analysis}
We visualized the anomaly curve (Fig. \ref{fig:anomaly_scores}) for the test video. It can be clearly observed that in the abnormal segments, our method outperforms AED-MAE~\cite{ristea2024self} and MA-PDM~\cite{zhou2025video}, producing anomaly scores that closely match the actual abnormal events. Moreover, in normal segments, our scores are more consistent and stable, resulting in a lower false alarm rate.

To verify the necessity of incorporating semantics, we also compared the baseline without semantics (Fig. \ref{fig:anomaly_scores}). It failed to effectively detect the abnormal event of a distant cyclist in the first half of the video, whereas the semantic significantly enhanced the perception of such local subtle anomalies.

Figure \ref{fig:error_maps} displays error maps between frame-centric predictions and GT, showing small errors under normal conditions but clear artifacts like color shifts and blurring during anomalies. It demonstrates the method's capability for accurate spatial anomaly localization without object-centric inputs. More experimental results can be found in the \textbf{supplementary files}.


\section{Conclusions}

In this paper, we propose a semi-supervised VAD framework named SIB-VAD. 
SIB-VAD introduces a novel feature filtering paradigm based on semantics-aware information bottleneck as an alternative to conventional memory modules, enabling a more direct and effective anomaly information filtering mechanism that adapts well to diverse scenarios.
Furthermore, by introducing a semantic branch that jointly utilizes appearance and motion to capture the high-level semantics in abnormal events, instead of relying solely on appearance and motion for insufficient local anomaly perception. 
Extensive experiments demonstrate that our method surpasses existing state-of-the-art approaches. 
We believe this work will provide new perspectives for future research.
\bibliographystyle{IEEEbib}
\bibliography{icme2026references}

\clearpage
\title{Video Anomaly Detection with Semantics-Aware Information Bottleneck (Appendix)}
\maketitle
\section{Implementation Details}
\subsection{Tri-Modal Fusion Encoder}
\begin{figure}
    \centering
    \includegraphics[width=\columnwidth]{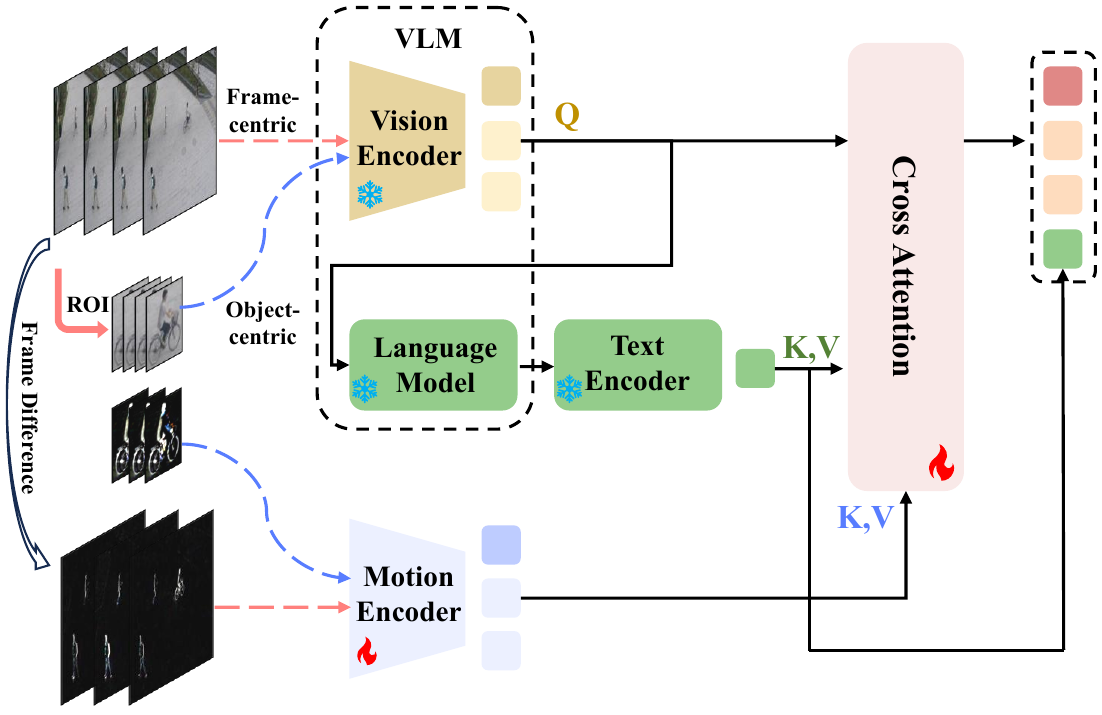}
    \caption{The architectural details of Tri-Modal Fusion Encoder.}
    \label{fig:tri}
\end{figure}

The detailed architecture of the tri‐modal fusion encoder is shown in Figure \ref{fig:tri}. For the vision appearance encoding branch, we directly apply the vision encoder from the VLM~\cite{chen2024expanding} to each input video frame. To match the VLM’s required input size, every frame is first resized to 448×448 pixels and then normalized. After passing through the encoder, each frame is mapped to a set of visual embeddings $\{\mathbf{h}^\text{cls}, \mathbf{h}^{1}, \mathbf{h}^{2}, \cdots, \mathbf{h}^{N_p}\}$, where ``cls'' denotes the class token, $N_p$ is the number of patches, and each token has dimensionality 1024.

For the text semantic encoding branch. The visual features produced above are projected and, together with a fixed prompt $\textbf{P}$, fed into a language model~\cite{yang2024qwen2} to generate a textual description of the video. To ensure steady outputs, we disable any sampling strategy and cap the generated text at a maximum length of 1024 tokens. The fixed prompt $\textbf{P}$ is: ``\emph{Describe the scene in the video segment, what objects are present, and what is each person doing?}''. We further employ a text encoder~\cite{gao2021simcse} to extract semantic features. This encoder uses contrastive learning to mitigate anisotropy in the semantic embeddings and to enhance the expressive power of the sentence representations. This step provides the foundational guarantee for all subsequent computations. Each descriptive text is encoded into a 768-dimensional semantic feature vector $\textbf{f}_{1:t}^{\text{text}}$.

For the motion dynamic encoding branch, we use frame differences as input. Since each frame-difference map can likewise be treated as an image, we train a ViT-based encoder to extract motion information and align its spatial dimensions with those of the visual features. The motion encoder consists of four encoding layers. Each frame-difference map is encoded into motion embeddings $\{\mathbf{m}^\text{cls}, \mathbf{m}^{1}, \mathbf{m}^{2}, \cdots, \mathbf{m}^{N_p}\}$, where ``cls'' denotes the class token and $N_p$ represents the number of patches.

We then fuse the outputs of the three branches to prepare for joint decoding. Concretely, we first project all modality features into the same dimensionality. 
To enable deeper interaction between triple modals, we use a cross-attention module to inject semantic and motion information into visual features.
Subsequently, the text semantic feature is inserted at the beginning of the token sequence, yielding the fused feature $\{\mathbf{f}_{1:t}^{\text{text}},\mathbf{f}^\text{cls}, \mathbf{f}^{1}, \mathbf{f}^{2}, \cdots, \mathbf{f}^{N_p}\}$.

\subsection{Sparse Feature Filtering Module}
\begin{figure}
    \centering
    \includegraphics[width=\columnwidth]{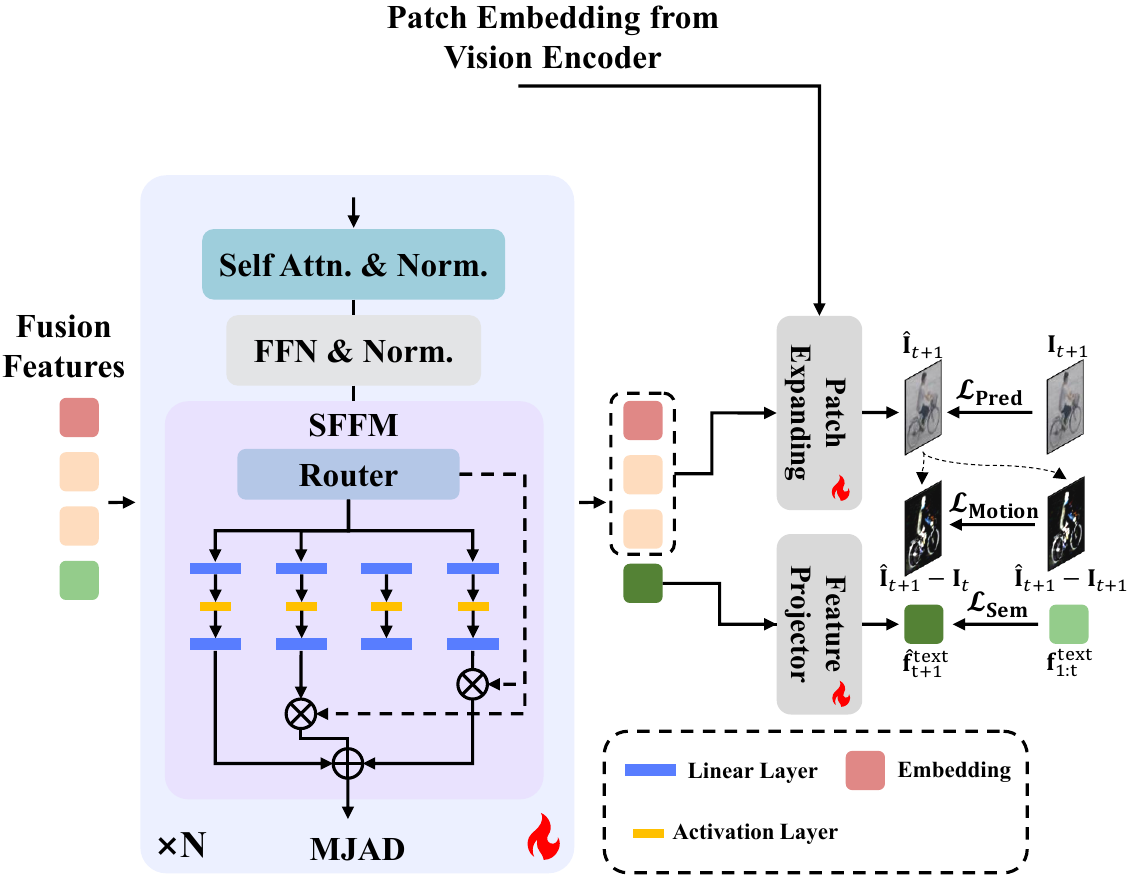}
    \caption{The architectural details of the Sparse Feature Filtering Module and Multimodal Joint Attention Decoder.}
    \label{fig:sffm_decoder}
\end{figure}

The detailed architecture of the Sparse Feature Filtering Module is shown in Figure \ref{fig:sffm_decoder}. In this module, the bottleneck filter serves as its basic unit. Concretely, each bottleneck filter is implemented as a bottleneck MLP consisting of two linear layers with a SiLU activation in between. The hidden-layer dimensionality is tuned per dataset to control the strength of feature filtering.

To further extend the capability of these bottleneck filters, we adopt an adaptive routing mechanism into the SFFM. By default, this architecture comprises 63 routing filters. For each input feature, a linear routing process computes its affinity to every filter; the top 7 filters (by adaptive gating score) are activated, and their outputs are combined via a softmax-weighted sum. Additionally, we include a shared filter to capture common knowledge and boost the specialization of the routed filters during feature filtering.

\subsection{Multimodal Joint Attention Decoder}
The detailed architecture of the decoder is shown in Figure \ref{fig:sffm_decoder}. We employ a ViT-based decoder to process the fused features. Its multi-head attention layers allow the fused features to interact with one another. Additionally, at each decoding layer, we insert a Sparse Feature Filtering Module to remove anomalous information. Because the model is trained exclusively on normal data, when anomalous behavior occurs, the combined effects of multi-head attention and feature filtering will distort and disrupt the feature representations, thereby amplifying prediction errors and making abnormal versus normal events easier to distinguish.

\subsection{Experimental Details}
Our approach follows a prediction-based paradigm, utilizing $t=4$ consecutive input frames to predict both the 5th frame and its semantic features. The frame error and semantic error jointly indicate anomaly confidence. We employ InternVL2.5-1B~\cite{chen2024expanding} as our pre-trained VLM and adopt SimCSE-$\text{BERT}_\text{base}$ (trained on NLI datasets)~\cite{gao2021simcse} as the pre-trained text encoder. For the frame-centered approach, we process entire frames directly. 

For the object-centric approach, we use the YOLOv11~\cite{khanam2024yolov11} object detector to extract bounding boxes, and then use each detected object independently as input. We take the highest anomaly score among all objects at the current moment as the frame-level confidence for the current frame. For frames where no objects are detected, we assign the lowest value as the anomaly score.

For model optimization, we utilize the Adam optimizer with a base learning rate of 0.0002, momentum hyperparameters $\beta_1=0.9$ and $\beta_2=0.999$, coupled with L2 weight decay of $1\times10^{-5}$. The depth of the motion encoder and the multimodal joint attention decoder is set to four layers. Training epochs are configured as (1000, 10, 5) for UCSD Ped2, CUHK Avenue, and ShanghaiTech, respectively. All frames are resized to $448\times448$ resolution. For the UCSD Ped2 and ShanghaiTech datasets, we set the reduction ratios of 32 and 16 for each bottleneck filter according to the characteristics of each dataset. For CUHK Avenue, whose frame is comparatively narrow and where foreground objects occupy a relatively large portion of the view, an excessive reduction ratio would over-filter vital information; therefore, we use a reduction ratio of 4/3. The loss weighting factors $\lambda_\text{1}, \lambda_\text{2}, \lambda_\text{3}$ are all set to 1 with $\lambda_\text{4}$ is 0.001. The anomaly score weights ($w_v$, $w_t$) for UCSD Ped2, CUHK Avenue, and ShanghaiTech datasets are set to (0.8, 0.2), (0.9, 0.1), (0.2, 0.8), respectively. All experiments are conducted on four NVIDIA GeForce RTX 4090 GPUs.

\begin{figure}
    \centering
    \includegraphics[width=\columnwidth]{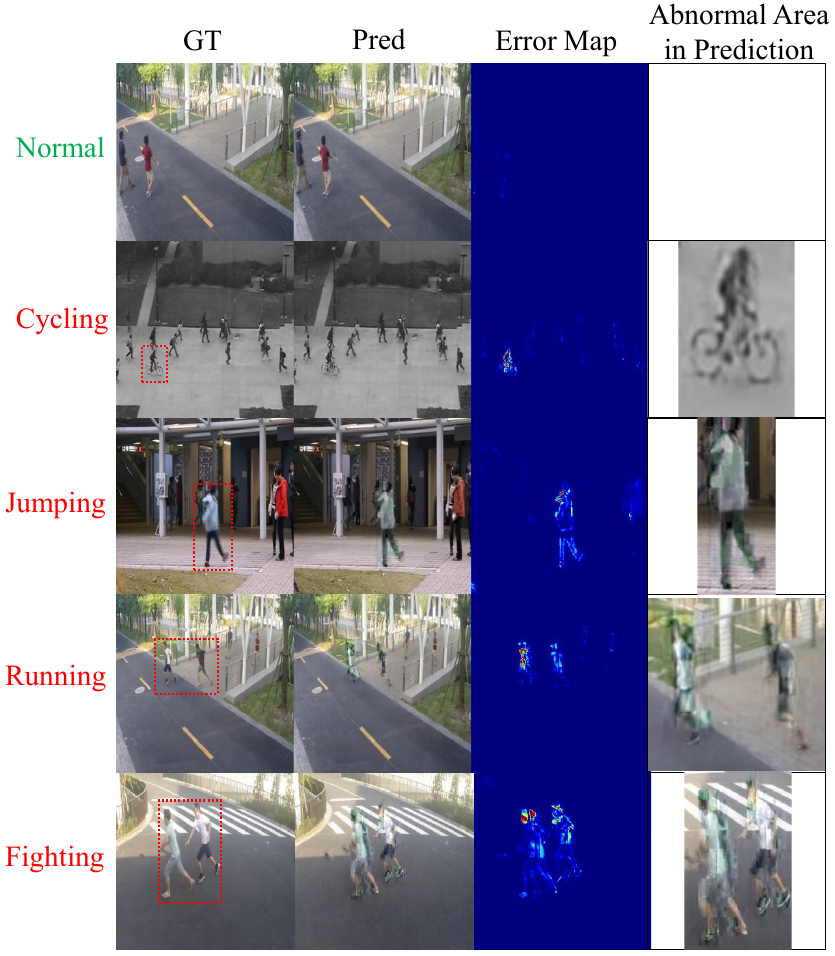}
    \caption{Examples of predictions and error maps for normal and different abnormal events.}
    \label{fig:appendix_error_maps}
\end{figure}

\section{More Qualitative Results}
\subsection{Anomaly Localization}
Figure \ref{fig:appendix_error_maps} shows additional visualizations of our prediction results alongside their corresponding error maps relative to the ground truth. Under normal conditions (first row), the ground truth and predicted frames are virtually indistinguishable, and the error map is completely blue, indicating negligible error and no detected anomalies. Below, we present several examples of predicted abnormal events—cycling, jumping, running, and fighting. In the case of cycling, the anomaly manifests in two ways: bicycles never appear in the Ped2 training set, so their appearance deviates from the normal pattern, and their rapid motion differs from typical pedestrian movement on sidewalks; in the predicted frames, noticeable missing pixels and blurring around both the bicycle and rider confirm these deviations. Similarly, for jumping, running, and fighting, the regions of interest in the predicted frames exhibit clear distortions—shape warping, blurring, and misalignment—which are further highlighted in the error maps. These examples demonstrate that our method is also effective in spatially localizing abnormal events.

\begin{figure}
    \centering
    \includegraphics[width=0.9\columnwidth]{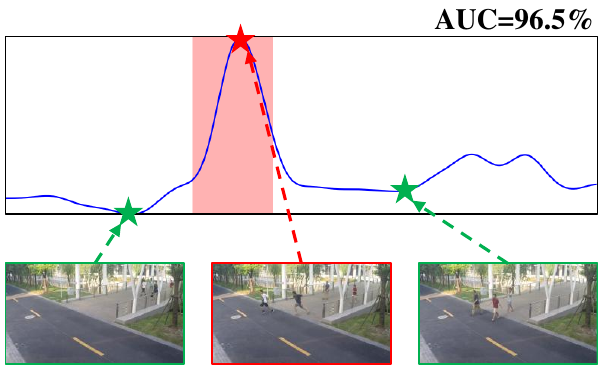}
    \caption{The anomaly score curve of the test video \textbf{05\_0018} from ShanghaiTech. The red intervals indicate anomaly segments, the blue curve represents the computed anomaly scores, and the images above show the corresponding abnormal events (marked in red) or normal events (marked in green).}
    \label{fig:05_0018}
\end{figure}

\begin{figure}
    \centering
    \includegraphics[width=0.9\columnwidth]{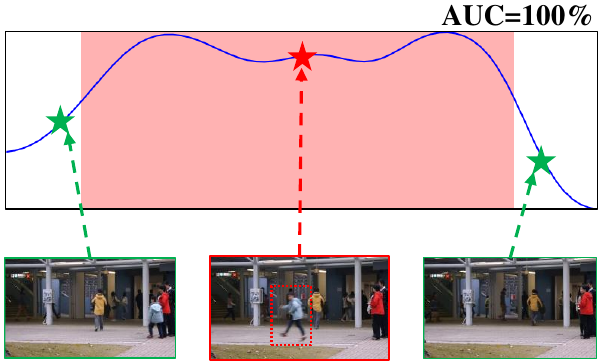}
    \caption{The anomaly score curve of the test video \textbf{21} from CUHK Avenue. The red intervals indicate anomaly segments, the blue curve represents the computed anomaly scores, and the images above show the corresponding abnormal events (marked in red) or normal events (marked in green).}
    \label{fig:21}
\end{figure}

\begin{figure}
    \centering
    \includegraphics[width=0.9\columnwidth]{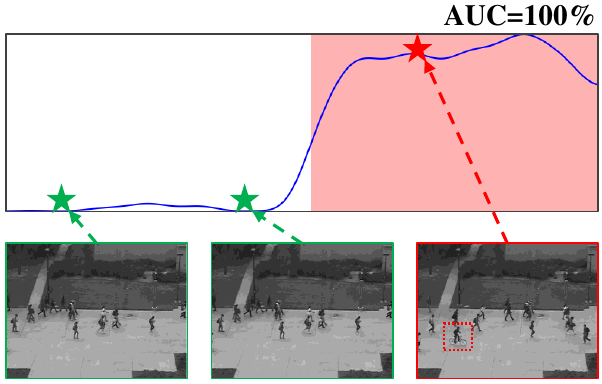}
    \caption{The anomaly score curve of the test video \textbf{Test002} from UCSD Ped2. The red intervals indicate anomaly segments, the blue curve represents the computed anomaly scores, and the images above show the corresponding abnormal events (marked in red) or normal events (marked in green).}
    \label{fig:Test002}
\end{figure}

\subsection{Anomaly Scoring}
We present more anomaly score curves to further demonstrate the effectiveness of our method. In Figure \ref{fig:05_0018}, we show the anomaly score curve for test video \textbf{05\_0018} from the ShanghaiTech dataset. In this video, an abnormal event occurs where pedestrians are seen chasing and playfully running on the road. Our method achieves an AUC score of 96.5\% on this test sample, indicating that it can accurately distinguish between anomalous and normal frames, successfully identifying the people who are running and chasing.

In Figure \ref{fig:21}, we present the anomaly score curve for test video \textbf{21} from the Avenue dataset. In this video, a child is seen jumping and playing on the sidewalk, which deviates from the normal pattern of pedestrians walking and is therefore considered abnormal behavior. As shown, our method achieves a perfect AUC score of 100\% on this video, demonstrating its ability to accurately detect abnormal events.

In Figure \ref{fig:Test002}, we show the anomaly score curve for test video \textbf{Test002} from the Ped2 dataset. In this video, an abnormal event is observed when a person is riding a bicycle in the pedestrian area. Our method is also able to perfectly identify and distinguish between normal and abnormal behaviors in this test sample, achieving an AUC of 100\%.


\begin{figure}
    \centering
    \includegraphics[width=0.9\columnwidth]{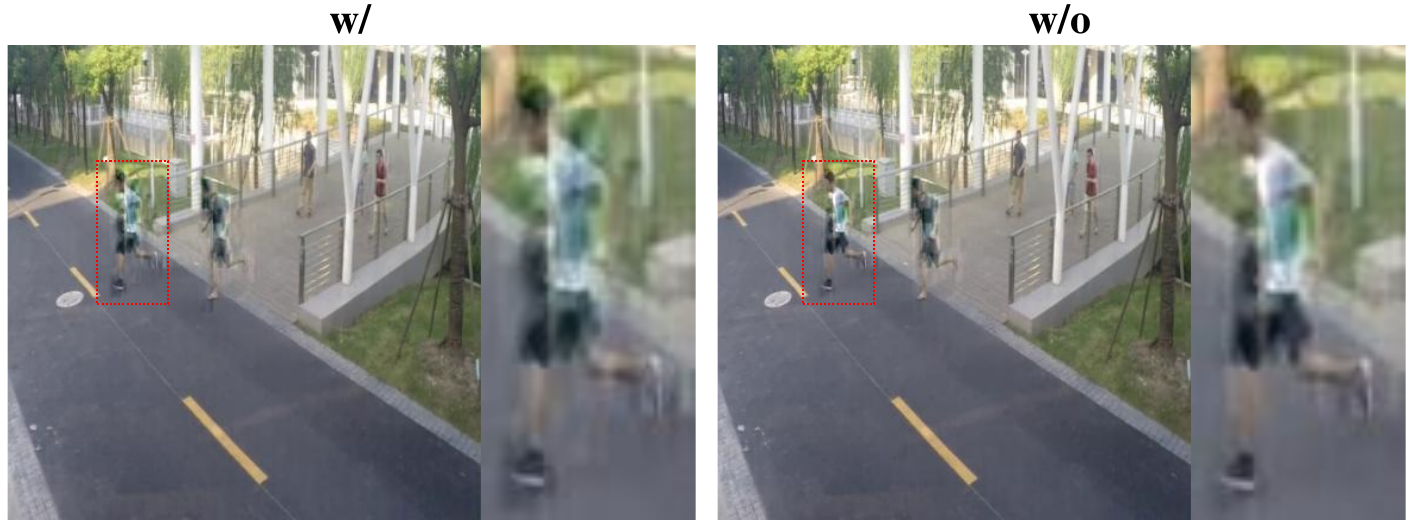}
    \caption{Ablation of Motion Frame-difference Contrastive Loss. The left side shows the prediction results with the motion frame-difference contrastive loss, while the right side shows the results without it. To enable a clearer comparison, one of the abnormal objects is enlarged.}
    \label{fig:motion_loss}
\end{figure}


\subsection{Ablation of Motion Frame-difference Contrastive Loss}
As shown in Figure \ref{fig:motion_loss}, we present the predicted frames with and without using the motion frame-difference contrastive loss. It can be observed that when abnormal behavior occurs, the predicted error for the abnormal object on the left is larger, with obvious pixel loss. In contrast, the prediction on the right retains more details around the edges of the abnormal object’s body. This indicates that incorporating the motion frame-difference contrastive loss can effectively improve anomaly detection performance and suppress the model from simply relying on skip connections in the vision encoder to mimic the previous frame for prediction.

\subsection{Compare with Memory Modules}
\begin{figure}
    \centering
    \includegraphics[width=0.9\columnwidth]{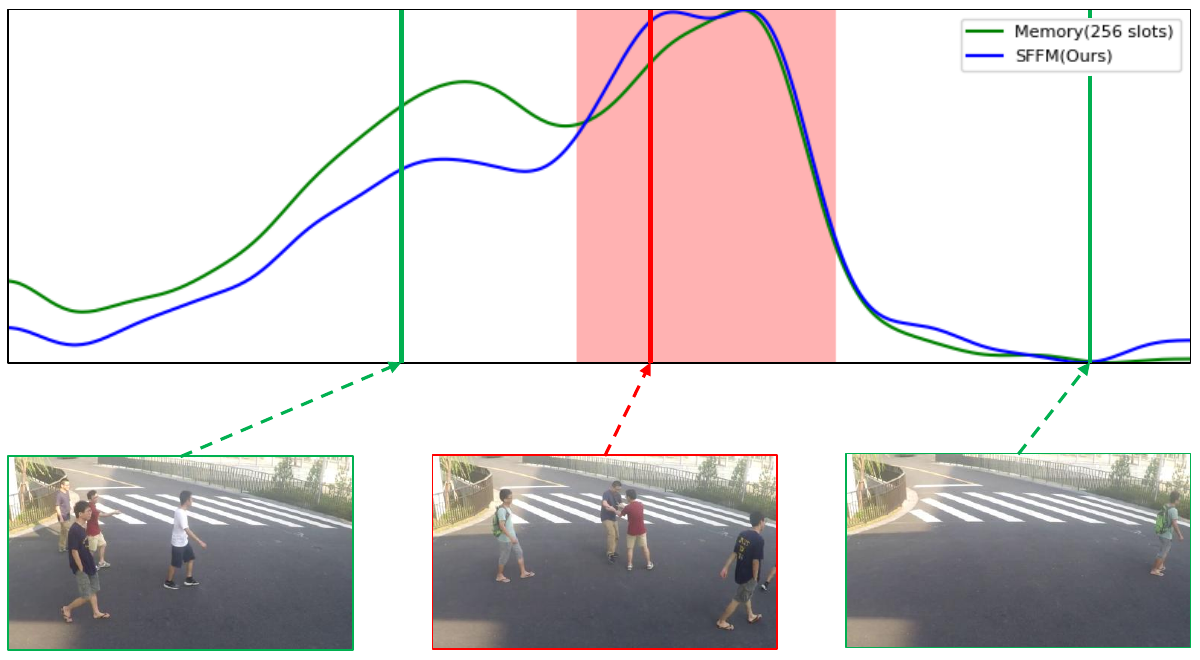}
    \caption{Comparative analysis with memory modules. The blue curve is obtained using our method, while the green curve is obtained by the memory-based method, which removes the SFFM and inserts a memory module of size 256 between the encoder and decoder. }
    \label{fig:cmp_w_mem}
\end{figure}

As shown in Figure \ref{fig:cmp_w_mem}, a visual comparison between our method and the memory module-based method is presented. The blue curve represents the results computed by our method. Consistent with the experiments in the main paper, we removed all SFFMs and inserted memory modules of size 256 in both the encoder and decoder for comparison, resulting in the green curve. From the figure, it can be observed that in the early stages of the video, when no abnormal behavior has yet occurred, the anomaly scores calculated by our method are significantly lower than those of the memory-based method, reducing the possibility of false positives. Conversely, when abnormal events occur, the anomaly scores computed by our method are higher than those of the memory-based approach. This validates that the proposed SFFM achieves better anomaly detection performance than the memory module.

\subsection{Failure Case}
\begin{figure}
    \centering
    \includegraphics[width=0.9\columnwidth]{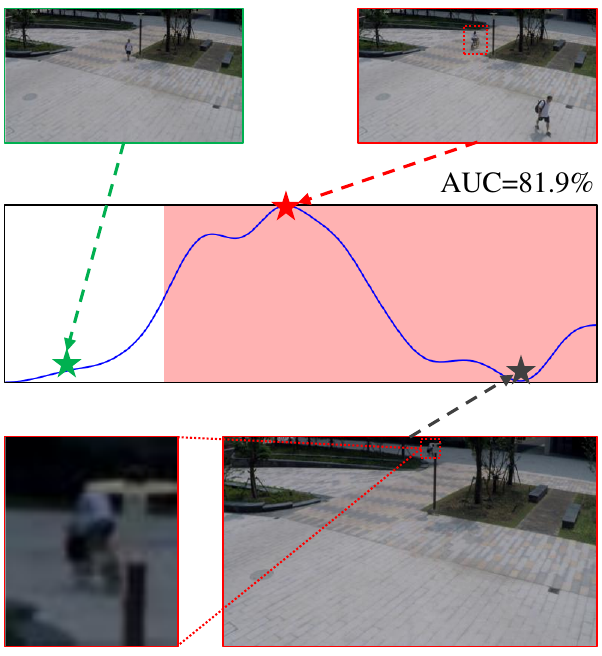}
    \caption{A failure case of our method.}
    \label{fig:failure_case}
\end{figure}

Although SIB-VAD is capable of accurately detecting most abnormal events, it occasionally fails to identify abnormal behaviors in some extreme scenarios. Figure \ref{fig:failure_case} shows a failure case of our method on test video \textbf{12\_0174} from the ShanghaiTech dataset. In this test video, the abnormal event involves a person riding a bicycle in a pedestrian area, as shown, although our method is able to compute high anomaly scores at the early stage of the event, indicating that the abnormal behavior is initially well detected. However, the anomaly scores gradually decrease as the rider moves farther away and becomes smaller in the frame, eventually dropping to zero. This reveals a significant misjudgment issue, and the AUC metric for this video only reaches 81.9\%.

This failure can be attributed to the fact that when the abnormal object is far away, it occupies a very small portion of the frame, which leads to two problems. First, the small size of the object results in very low frame prediction errors, as the static background dominates and averages out the prediction errors. Second, when the object is small, the VLM struggles to capture accurate semantic information about the abnormal object. It will also lead to a low semantic prediction error. Overall, even if we introduce semantics, it can, to some extent, solve the ability of appearance and motion to perceive local anomalies. However, when facing overly small foreground goals and the semantics cannot be well captured either, our multimodal framework will fail.

We conjecture that further incorporating multi-scale temporal semantic information integration can help alleviate this issue. Specifically, in addition to the existing short-term window of 4 consecutive frames, temporal windows with larger time steps—such as 8 frames, 16 frames, 32 frames, and so on—can be constructed.  Video descriptions can then be extracted from these different levels of frame sequences, and the semantic features from each level can be fused to capture the evolution of behavior over a longer temporal range within the scene.   Furthermore, incorporating object-tracking techniques and integrating semantic information may further improve the accuracy of anomaly detection.     However, the added complexity and computational overhead brought by the integration of multi-scale temporal semantics and object tracking pose significant challenges.   Besides, how to effectively fuse semantic features from different levels remains a key challenge. We leave these considerations for future research.

\section{Limitations}
The goal of this paper is to achieve frame-level video anomaly detection. By introducing an information bottleneck network with semantic awareness, we obtain improved performance in frame-level anomaly detection. However, this approach still has certain limitations when dealing with small objects, because both frame-centric methods and object-centric methods that incorporate additional object-detection priors remain insufficiently sensitive to distant and small objects. Moreover, identifying the types of anomalies is also crucial for video anomaly detection systems. Future work needs to further address challenges such as detecting spatially small anomalies and distinguishing between different categories of abnormal behaviors.

\section{Broader Impact}
SIB-VAD is a novel Video Anomaly Detection framework. By introducing semantic modalities and sparse feature filtering mechanisms, a high-performance frame-centric video anomaly detection capability has been achieved, bringing more diverse perspectives and contributions to the VAD field. This method is conducive to improving the system's energy efficiency in scenarios such as monitoring security and network content review, efficiently detecting and identifying abnormal events in videos, and bringing beneficial social impacts.

However, like many video analysis technologies, SIB-VAD may raise concerns regarding privacy and potential misuse. In surveillance or content monitoring scenarios, the deployment of automated anomaly detection systems could lead to excessive data collection or surveillance overreach, especially without appropriate safeguards or oversight. We are committed to the ethical development and responsible deployment of our technology and will adhere to relevant data protection laws and industry best practices. 

\end{document}